# Machine Learning-based Nutrient Application's Timeline Recommendation for Smart Agriculture: A Large-Scale Data Mining Approach

Usama Ikhlaq[a], Tahar Kechadi[a]

[a] Department of Computer Science, University College Dublin (UCD)

Dublin, Republic of Ireland

usama.ikhlaq@ucdconnect.ie, tahar.kechadi@ucd.ie

*Abstract*— This study addresses the vital role of data analytics in monitoring fertiliser applications in crop cultivation. Inaccurate fertiliser application decisions can lead to costly consequences, hinder food production, and cause environmental harm. We propose a solution to predict nutrient application by determining required fertiliser quantities for an entire season. The proposed solution recommends adjusting fertiliser amounts based on weather conditions and soil characteristics to promote cost-effective and environmentally friendly agriculture. The collected dataset is high-dimensional and heterogeneous. Our research examines large-scale heterogeneous datasets in the context of the decision-making process, encompassing data collection and analysis. We also study the impact of fertiliser applications combined with weather data on crop yield, using the winter wheat crop as a case study. By understanding local contextual and geographic factors, we aspire to stabilise or even reduce the demand for agricultural nutrients while enhancing crop development. The proposed approach is proven to be efficient and scalable, as it is validated using a real-world and large dataset.

## I. Introduction

Digital agriculture creates autonomous procedures that can be tailored to specific needs. Data is the backbone of digital agriculture and is mostly sensor data (e.g., in-field and remote sensors), farm machinery (e.g., tractors), and manual inputs. The inter-relationship between agriculture data features (dimensions) is the core factor in determining the right decision for farming practices, i.e., by monitoring the climate, soil and crop, the decision-support system can predict the application of fertiliser [1]. This can be done for a particular arable field in the form of predictions by taking suggestions from the system [2].

There are two categories of fertilisers: organic and (non-organic) chemical fertilisers [3]. These categories have different applications and timelines for a given field. Usually, chemical fertilisers consist of primary and secondary nutrients. Organic fertiliser comes from the dung of animals and plants and is usually used to increase soil fertility [4].

Many researchers are working on fertiliser optimisation issues in agriculture. Especially after the inflation in natural gas prices in 2001 and recently in 2022, nitrogen (N) fertiliser prices reached nearly double, and as a result, the world use efficiency for cereals gain production was only 33% in 2001, which was a significant decrease [5]. There is also an estimation that in 1999, the unaccounted 67% in Nitrogen represented a 15.9 billion annual loss of N fertiliser [6]. This estimation, due to the shortage of natural gas in 2002, was equal to an unaccounted 67% more than 20 billion annually [5]. As per the world consumption of Nitrogen by the Food and Agriculture Organization (FAO), the world demand for N fertiliser was 105,148 metric tons in 2016 (FAO, 2016), 107,424 metric tons in 2019 (FAO, 2019) and 111,591 metric tons in 2022 (FAO 2022). As can be seen over time, there continues to be significant and increasing demand for N fertiliser.

To achieve the fertiliser application optimisation objective, various factors need to be considered, including weather conditions i.e., sunlight, rainfall, and wind. It is also important to understand which nutrients in fertiliser are needed for crop growth and in which combination [7]. It is also more important to know which growth stage the crop is in relation to the seeding date. By considering the above scenario the problem becomes more complex to determine which nutrient is needed and in how many applications and when to apply and in which quantity [8].

Moreover, the decision-based algorithms can do more, i.e., as they can suggest the fertiliser applications according to the soil type and the area of the farm, they are smarter in the calculation. For this purpose, I have used the decision trees as the environmental and ground-level attributes change all the time and need to take the final decision on the bases of prior multiple decisions [9]. To solve the problem, multiple decisions need to be taken and for this purpose, the Random Forest algorithm can work based on multiple decision trees and conclude all the decisions on the bases of the voting system to formulate the final decision. This works so well when working with randomness in attributes and voting with randomness. So, for this research, the Random Forest algorithm has been chosen to solve the problem [10][11].

## II. Machine Learning based Approaches

Predictions in the agriculture domain are equivalent to coping with randomness. Consider a real environment along with the continued climate change like temperature, rainfall, cloud coverage, air pressure, humidity, and soil conditions along with cropping seasons. Due to these variable factors, crop production is mostly affected. To achieve accurate predictions near to reality, better environmental and Agri-decision-based systems are required to be developed and implemented. To cope with this problem, we have reviewed several Machine Learning (ML) algorithms to analyse their learning patterns, accuracy, and combinations.

In [12]–[14], the authors worked on Crop yield prediction, including winter wheat forecasting. They used the ML regression-based analyses algorithms, which is a traditional way of estimating crop yield with primarily trusted models for crops. Recently, it was reported that these models can replicate crop growth and the key factors with detail and can run on various levels [16, 15]. Regression modelling experiments often rely on computationally rigorous processes requiring fine-grained data, such as daily weather changes and fertilizer applications, which limit their large-scale applicability [17, 18]. On the other hand, statistical models are only effective for addressing homogeneous data with smaller dimensionality and cannot be easily extended to different problem sets (higher dimensionality, etc.) [14, 19].

In recent years, machine learning (ML) models have emerged as viable alternatives to traditional statistical models [20]. ML models, which operate based on weighted importance rather than probabilities, can effectively replace statistical models. Various ML techniques, such as Random Forest (RF), have demonstrated their ability to handle noisy data and interpret nonlinear relationships [21, 22]. Moreover, ML models have been increasingly employed for crop classification and yield prediction in recent years [23, 24].

Deep learning (DL) models, a subset of ML, have shown promise in various fields, including image and speech recognition [26, 27, 28]. DL has been used for data-driven Earth System Science to perform complex ML tasks [25]. However, only a few studies have applied DL for fertiliser prediction and yield estimation [25]. Moreover, data-driven approaches based on weather conditions have not been thoroughly investigated compared to regression techniques in the case of fertiliser applications.

There have been significant advances in Random Forest (RF) developments by combining decision tree prediction with classification, regression, and ensemble techniques [10]. Furthermore, RF has improved by introducing error splitting for each node yield, making the algorithm more insensitive to noise [29]. This internal assessment of variable importance applies to regression as well. In [30, 31, 32], the authors investigated the RF algorithm's growth and the selection of training sets in multiple studies. The "random subspace method for constructing decision forests," which uses feature selection for each tree's growth, has been reported to be efficient [33]. [11] gives more details about RF classification and regression.

### III. MULTIVARIATE REGRESSION TREES

Multivariate Regression Trees (MRTs), also known as multi-target regression trees, multi-objective regression trees, or multi-output regression trees, can simultaneously predict multiple continuous target values [34]. The two main advantages of multi-target regression trees over separate construction of regression trees for each target are smaller size and better recognition of dependencies among target variables.

MRTs have received much attention, with notable work by De'ath, who proposed the first approach to multivariate regression [35]. MRTs were developed along the same lines as CART, initially including all instances in the root node, then iteratively finding the best possible split until a predefined stopping criterion is met. The main difference from CART lies in the redefinition of node impurity, determined by the sum of squared errors in response to multiple targets.

In [36], the authors highlighted that MRTs inherit several attributes of univariate regression trees, such as ease of interpretation and construction, robustness against noise for predictor variables, automatic detection of interactions among features, and effective handling and missing feature variable information reduction.

### IV. LARGE-SCALE DATA: TIMELINE CONSTRUCTION

To prepare the data for the ML model, we were faced with a huge challenge of dealing with significant missing values, noise, and outliers. Moreover, the data is very large and heterogeneous. For predicting nutrient application timelines and varying fertiliser quantities based on weather conditions and growth stages, we extracted sample data that consists of about 3000 fields. We pre-process the data by developing efficient algorithms and methods to deal with missing values, dimensionality reduction, smoothing noise, and outliers, and so on. To address potential ML model development issues, such as biases or over-fitting, we devised a strategy that involved subdividing the datasets into smaller subsets and constructing timelines for each nutrient's application.

The goal was to emulate farmers' practices by gathering all necessary attributes in a timeline format for each nutrient's application. This includes factors, such as weather conditions during fertiliser application, time gaps between applications, and any special conditions related to individual nutrients. By isolating these sub-factors, we were able to connect them and predict nutrient application timelines effectively.

We developed a timeline construction method that incorporates the subdivided data, which focused primarily on nitrogen but could be adapted for other nutrients. The process consisted of the following steps:

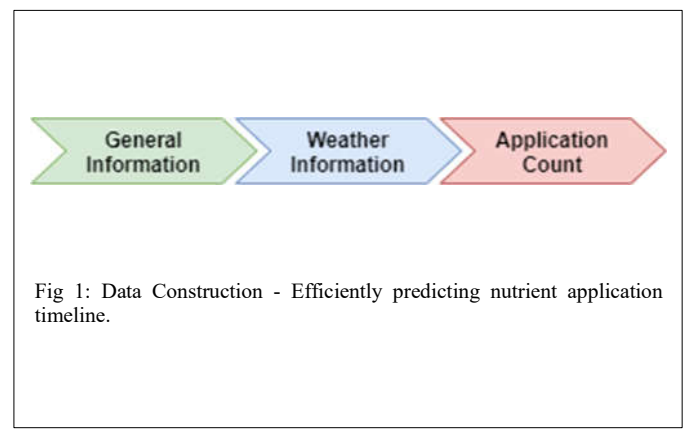

Fig 1: Data Construction - Efficiently predicting nutrient application timeline.

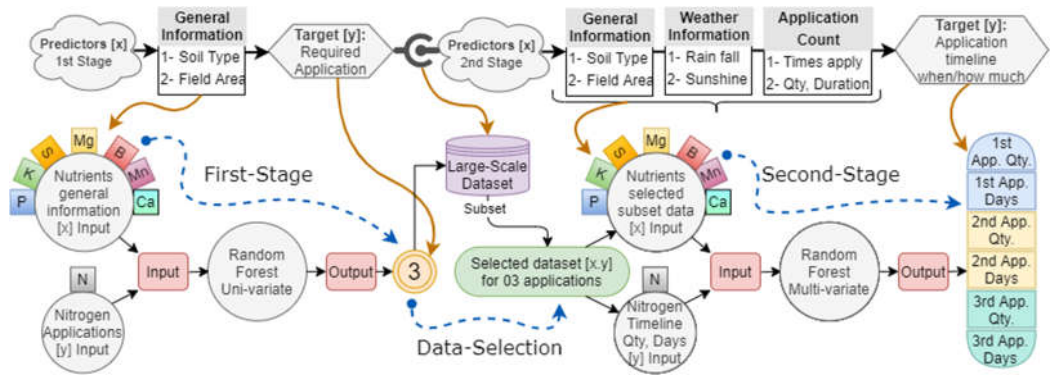

Fig 2: Simplified Model - A brief illustration of a two-stage model predicting nitrogen timelines, applicable to other nutrients.

## V. SYSTEM ARCHITECTURE

The system architecture, illustrated in Fig. 3, comprises three parts: data management, Random Forest regression, and multi-output regression using a multivariate approach. Raw data, including weather, fertilizer, and nutrient application timelines, is cleaned, and processed using techniques like nutrient dictionaries and feature extraction. This data is prepared for machine learning through encoding and conversion of categorical attributes, such as soil type. Fertilizer applications are separated, and data pipelines are created for each. Data is then divided for univariate and multivariate approaches, enabling prediction of multiple outputs. The two models are pipelined, with the output of the first model serving as input for the second. Further component details follow in subsequent sections.

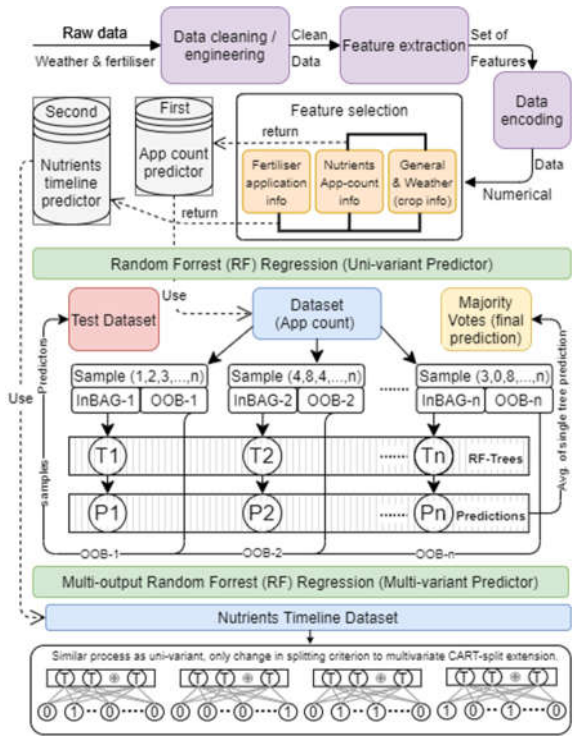

Fig. 3: System Architecture Overview - Illustrating the three-part division of the system, including data management, Random Forest regression model, and multi-output Random Forest regression using a multivariate approach.

## VI. SYSTEMATIC MODEL

We developed a two-stage simplified systematic model (Fig 2) specifically for predicting nitrogen applications. The first stage determines the required number of nitrogen applications, while the second stage predicts the quantity and days for each application in a timeline format. A subset data selection process from the main data occurs between model stages 1 and 2.

In the first stage, the model takes predictors as input, which are the application predictors for Nitrogen (here we are predicting nitrogen application count with the help of other nutrients application) and other nutrients such as phosphorus (P), potassium (K), sulphur (S), magnesium (Mg), boron (B), manganese (Mn), and calcium (Ca). The target variable (Y) is the number of nitrogen applications. By training the model with these inputs, it can provide an output indicating the required number of applications of nitrogen, for example, 3 applications.

Once the first stage determines that three applications are required, the relevant sub-data for these three applications is selected as input for the second stage. The input attributes consist of weather predictors and application-specific nutrient values for each of the 3 applications (e.g., N3, P3, K3, S3, Mg3, B3, Mn3, Ca3). In the second stage, the target variables (Y) for nitrogen are the quantity and the number of days from the seeding date for each application, i.e. The predicted days could be 115, 185 and 220 respectively from the seeding dates.

By employing this two-stage model, we can efficiently predict not only the number of nitrogen applications required but also the specific quantity and days for each application. This approach allows for more precise nitrogen management.

In this research, we first used a simplified model to predict the number of nutrient applications for nitrogen. The simple model helped to provide a basic understanding of the methodology and its functionality. We explained how the model predicts nitrogen application counts using X-Y predictors and target variables.

After understanding the basic model, we delved into a more complex real-world scenario where the model was applied to all nutrients. We briefly described the two-stage model architecture and the process of generating input data timelines based on the predicted number of applications for each nutrient. This advanced version Fig. 4 demonstrated the model's complexity and effectiveness in predicting nutrient application timelines across all nutrients, providing a comprehensive understanding of the methodology.

The results obtained from our research are based on the complex model designed for all nutrients (N, P, K, S, Mg, B, MN, Ca). This advanced version of the model allows for a more accurate and comprehensive prediction of nutrient application timelines and considers the various factors and interactions between the different nutrients. The simple model served as an introductory step to help understand the basic concepts, while the results were derived from the more sophisticated and realistic complex model.

Based on the given data, the maximum nutrient application count (app max count) has been determined in TABLE 3 for both the available data and the predicted model on the test data

The model's results generated a timeline for each nutrient based on the required number of applications, as elaborated in the following Evaluation section.

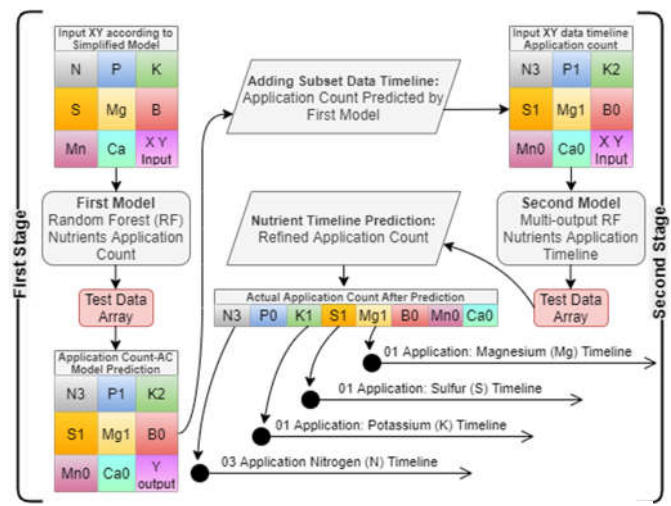

Fig. 4: Advanced Two-Stage Model - Illustrating the architecture for predicting application timelines, quantities, and days for all nutrients.

## VII. EVALUATION

We evaluated the performance of our models using the accuracy of the train and test sets, along with cross-validation to ensure that the models were not overfitting and could generalize well to new data. To further assess the robustness of our models, we calculated the standard deviation of the cross-validation results in TABLE 1 and TABLE 2 respectively. A higher standard deviation indicates that the model's performance is weaker, and the data may not fit well. Conversely, a lower score indicates a more robust model. We trained each nutrient separately using a similar model configuration, albeit with different datasets and features.

### A. Nutrients Application Count

The first stage of model (Fig 2) predicts nutrient application counts for each nutrient during the cropping season. Using general crop and farm information, it requires less data and training time than the second model. With a focus on nitrogen, we employed 10-fold cross-validation for evaluation.

The model achieved training accuracy results between 0.97% and 0.99%, and test accuracy results between 0.74% and 0.94% for each nutrient. However, fewer applications were observed for the last three nutrients (Boron, Manganese, and Calcium), as they are rarely required by crops. This led to instances with no application, affecting testing and cross-validation results. Manganese had 0 accuracy during testing but 0.55% accuracy during 10-fold cross-validation. Standard deviation results were better for the first five nutrients and average for the last three, with potential for improvement by increasing the amount of data.

TABLE 1: ACCURACY FOR THE FIRST MODEL NUTRIENTS APPLICATION COUNT PREDICTION.

| First Model Application Count | Predicted Maximum Application | Train Accuracy | Test Accuracy | Cross Validation Fold | Cross Validation | CV Standard Deviation |
|---|---|---|---|---|---|---|
| Nitrogen | 7 | 0.982 | 0.864 | 10 | 0.889 | 0.0138 |
| Phosphorus | 6 | 0.970 | 0.742 | 10 | 0.781 | 0.0429 |
| Potassium | 5 | 0.977 | 0.836 | 10 | 0.841 | 0.0105 |
| Sulphur | 6 | 0.989 | 0.944 | 10 | 0.937 | 0.0175 |
| Magnesium | 5 | 0.976 | 0.883 | 10 | 0.843 | 0.0800 |
| Boron | 1 | 0.993 | 0.963 | 10 | 0.559 | 0.390 |
| Manganese | 0 | 0.996 | 0.0 | 10 | 0.555 | 0.464 |
| Calcium | 1 | 0.987 | 0.854 | 10 | 0.924 | 0.059 |

### B. Nutrients Application Timeline

The second stage of model (Fig. 4) predicts the nutrient application timeline, determining the timing of crop fertilization for each nutrient during the cropping season. This stage used a comprehensive dataset, focusing on each nutrient's accuracy and validation for reliable predictions.

Promising results were observed for some nutrients (e.g., nitrogen, magnesium, and boron), with training accuracy ranging from 0.935 to 0.975 and test accuracy from 0.712 to 0.995. However, no application was added for manganese, and the model's accuracy for phosphorus and calcium was relatively lower. Cross-validation results varied for each nutrient, with standard deviation scores ranging from 0.0134 to 0.797.

Both stages did not predict any manganese application, indicating no requirement in the given conditions. The second stage model's accuracy results were lower than the first stage, likely due to the complexity of predicting the nutrient application timeline. Nevertheless, the model provides valuable insights for optimizing fertilization practices and achieving better crop yields.

TABLE 2: ACCURACY FOR THE SECOND MODEL NUTRIENTS APPLICATION TIMELINE PREDICTION.

| Second Model Application Timeline | Predicted Maximum Application | Train Accuracy | Test Accuracy | Cross Validation Fold | Cross Validation | CV Standard Deviation |
|---|---|---|---|---|---|---|
| Nitrogen | 7 | 0.975 | 0.840 | 10 | 0.811 | 0.0103 |
| Phosphorus | 6 | 0.948 | 0.811 | 10 | 0.494 | 0.797 |
| Potassium | 5 | 0.935 | 0.712 | 3 | 0.449 | 0.144 |
| Sulphur | 6 | 0.964 | 0.887 | 4 | 0.817 | 0.0134 |
| Magnesium | 5 | 0.987 | 0.972 | 10 | 0.802 | 0.116 |
| Boron | 1 | 0.989 | 0.995 | 5 | 0.755 | 0.718 |
| Manganese | 0 | 0 | 0 | 0 | 0 | 0 |
| Calcium | 1 | 0.978 | 0.777 | 10 | 0.802 | 0.051 |

### C. Crop Nutrient Application Schedule

The results present a learning-based crop nutrient application schedule specifically for winter wheat crops, developed using best practices. The model takes an input test data array and extracts the timeline for each required nutrient based on crop needs. As detailed in Section VI, the first stage model generates nutrient application predictions, while the second stage model refines these predictions using a subset of the dataset. This process streamlines nutrient application requirements according to the given large-scale dataset. Table 3 displays the actual predicted application values from the first stage model and the refined and actual application recommendations from the second stage model. The refined results serve as the basis for generating the following nutrient timelines.

TABLE 3: COMPARISON OF NUTRIENT APPLICATION PREDICTIONS BETWEEN THE FIRST MODEL AND THE REFINED

| No. | Nutrients | First Model Application Prediction | Second Model Refine Outcome |
|---|---|---|---|
| 1 | Nitrogen | 3 | 3 |
| 2 | Phosphorus | 1 | 0 |
| 3 | Potassium | 2 | 1 |
| 4 | Sulphur | 1 | 1 |
| 5 | Magnesium | 1 | 1 |
| 6 | Boron | 0 | 0 |
| 7 | Manganese | 0 | 0 |
| 8 | Calcium | 0 | 0 |

### D. Timeline-Driven Outcomes

In the timeline-driven outcomes section, we present graphical representations of nutrient application schedules for Nitrogen, Potassium, Sulphur, and Magnesium, using consistent unit indicators and colours to differentiate between the predicted results for each nutrient. It is worth mentioning that the model did not recommend any application requirement for Phosphorus, Boron, Manganese, and Calcium according to the given features, and different numbers of recommendations were given for the rest of the nutrients. The timelines showcase the quantities of nutrients and yield, as well as the corresponding application days for each nutrient throughout the crop's growing season.

- n_kg/ha [green round pointer]: Nitrogen - Kilogram per hectare.
- n_app_days [red diamond pointer]: Application days for nitrogen from the day of seeding,
- total_n/ha [sky-blue triangle pointer]: Total nitrogen used throughout the cropping season.
- total_yield/ha [blue triangle pointer]: Total yield that can be expected after applying nutrient applications at the time of harvesting, measured in tons.
- n_qty [bars indicator]: Nitrogen - Kilogram per hectare.
- n_app_days_sum [bars indicator]: Application days for nitrogen from the day of seeding.
- total_n/ha_qty [bars indicator in sky-blue]: Total nitrogen used throughout the cropping season.
- total_yield/ha_ton [bars indicator in blue]: Total yield that can be expected after applying nutrient applications at the time of harvesting, measured in tons.

*1) Nitrogen Application Timeline:* shows three recommended applications at approximately 41, 58, and 78 Kg/ha of farmland, with the first application at around 115 days after seeding, the second at around 185 days, and the third at around 220 days. The predicted total yield and total nitrogen required are 9.9 tons/ha and 177 Kg/ha, respectively.

*2) Potassium Application Timeline:* suggests only one application of approximately 24 Kg/ha, with a total recommended application of 25 Kg/ha. The predicted crop yield is similar to that for nitrogen, at 9.8 tons/ha.

*3) Sulphur Application Timeline:* recommends one application at around 82 Kg/ha, to be applied at 117 days from the seeding date, with a predicted yield of 9.8 tons/ha. However, the total application predicted is 142 Kg/ha, which is higher than the single application prediction, warranting further investigation in future research.

*4) Magnesium Application Timeline:* indicates a single application of approximately 15 Kg/ha, with a total predicted application of 20.5 Kg/ha. The predicted yield is consistent across all nutrient applications.

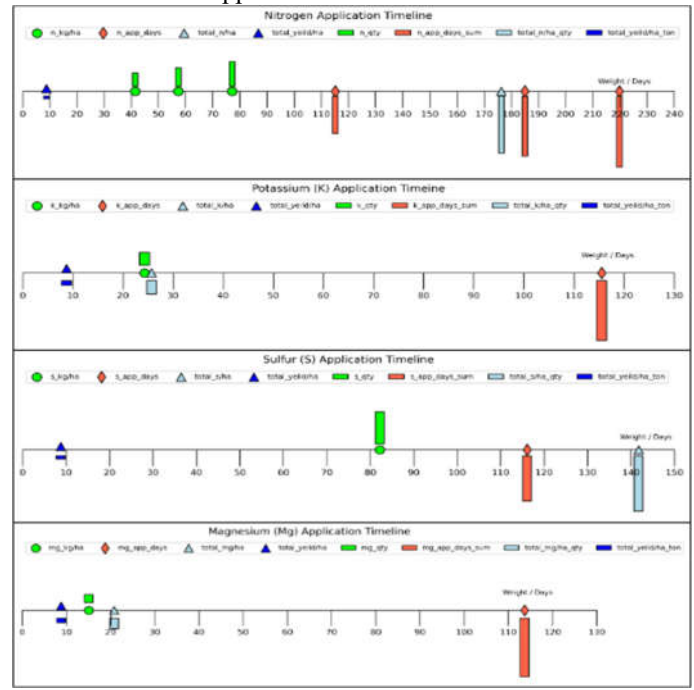

Fig 5: Application Timelines -Nitrogen, Potassium, Sulphur, and Magnesium, showcasing recommended nutrient quantities and application days.

### VIII. CONCLUSION

In conclusion, our research presents a novel two-stage ML method addressing large-scale dataset heterogeneity by selecting subsets for each nutrient's timeline. We predicted nutrient application timelines, including application count, precise days, and necessary quantities for the entire cropping season, while estimating crop yield based on recommended nutrient applications in fertilizers.

Our primary contributions include an innovative technique to handle large-scale datasets with a systematic two-stage model and the effective prediction of nutrient application timelines, applicable to medium and large-scale businesses.

For future work, we aim to reduce the feature set while maintaining or enhancing the model's accuracy and refine the prediction process by estimating combined nutrient applications, considering plants may require multiple nutrients on the same day, as shown in Fig 5.

By emphasizing our unique two-stage simplified and advanced systematic model and our ground-breaking approach to utilizing expansive data for selecting subset datasets in line with nutrient application timeline demands, we outline future endeavours while accentuating our technique's practical applicability and intricacy. Our work presents a notable contribution to the computer science domain, addressing agricultural fertilizer application challenges and large-scale data administration.

ACKNOWLEDGMENT

CONSUS is funded under Science Foundation Ireland's Strategic Partnerships Programme (16/SPP/3296) and is co-funded by Origin Enterprises Plc.